\documentclass{article}
\usepackage{ijcai20}

\usepackage{times}

\usepackage{soul}
\usepackage{url}

\usepackage[utf8]{inputenc}
\usepackage[small]{caption}
\usepackage{graphicx}
\usepackage{amsmath}
\usepackage{amssymb}
\usepackage{booktabs}
\usepackage{microtype}
\usepackage{amsthm}
\usepackage{amsmath}
\usepackage{inconsolata}
\usepackage{listings}
\usepackage{xcolor}
\usepackage{lipsum}
\usepackage[bitstream-charter]{mathdesign}
\usepackage{makecell}

\input{tikz_staff}

\newcommand{\nesy}{NeSy}

\definecolor{verydarkgreen}{rgb}{0.0, 0.5, 0.0}

\theoremstyle{definition}

\title{From Statistical Relational to Neuro-Symbolic Artificial Intelligence}

\author{
Luc De Raedt\and
Sebastijan Dumančić\and
Robin Manhaeve\And
Giuseppe Marra
\affiliations
KU Leuven, Department of Computer Science\\
\emails
\{firstname.lastname\}@kuleuven.be,
}
\begin{document}

\maketitle

\begin{abstract}
Neuro-symbolic and statistical relational artificial intelligence
both integrate frameworks for learning with logical reasoning.  This survey identifies several 
parallels across seven different dimensions between these two fields. These cannot only be used to characterize and position neuro-symbolic artificial intelligence approaches but also to identify a number of directions for further research. 


\end{abstract}

\section{Introduction}
The integration of learning and reasoning is one of the key  challenges in  artificial intelligence and machine learning today, and various communities have been addressing it.
That is especially true for the field of neuro-symbolic computation (NeSy) \cite{besold2017neural,garcez2019neural}, where the goal is  to incorporate symbolic reasoning into neural networks.
NeSy already has a long tradition, and it has recently attracted a lot of attention from various communities (cf. e.g., the  keynotes of Yoshua Bengio and Henry Kautz on this topic at AAAI 2020). 
Many approaches to \nesy{} aim at extending neural networks with logical reasoning. 

Another domain that has a rich tradition in integrating learning and reasoning is that of statistical relational learning and artificial intelligence (StarAI) \cite{Getoor07:book,raedt2016statistical}. 
But rather than focusing on how to integrate logic and neural networks, it is centred around the question of how to integrate logic with probabilistic graphical models.
Despite the common interest in combining logic or symbolic reasoning with a basic paradigm for learning, i.e., probabilistic graphical models or neural networks, it is surprising that there are not more  interactions between these two fields.  

This discrepancy is the key motivation behind this survey: it aims at pointing out the similarities  between these two endeavours and in this way stimulate more cross-fertilization. 
In doing so, we start from the literature on StarAI because, arguably, there is more consensus on what the key concepts, challenges and issues are in StarAI than in \nesy{} (cf. the number of tutorials and textbooks on related topics such as \cite{russell2015unifying,raedt2016statistical} but see also \cite{besold2017neural,garcez2019neural}).  
It turns out that essentially the same issues and techniques that arise in StarAI have to be addressed in \nesy{}  as well. 
The key contribution of this survey is {\em that we identify a set of seven dimensions that these fields have in common and that can be used to categorize both StarAI and \nesy{} approaches}.  
These seven dimensions are concerned with (1)  directed vs undirected models, (2)  grounding vs proof based inference, (3) integrating logic with probability and/or neural computation, (4) logical semantics,  (5) learning parameters or structure, (6) representing entities as symbols or sub-symbols, and, (7) the type of logic used.
We provide evidence for our claim by positioning a wide variety of StarAI and \nesy{} systems along these dimensions and pointing out analogies between them. 
This, in turn, allows us to identify interesting opportunities for further research, by looking at areas across the dimensions that have not seen much work yet.
Of course, there are also important differences between StarAI and \nesy{,} the most important one being that the former operates more at the symbolic level, lending itself naturally to explainable AI, while the latter operates more at the sub-symbolic level, lending itself more naturally for computer vision and natural language processing.

Unlike some other recent surveys or perspectives on neuro-symbolic computation \cite{besold2017neural,garcez2019neural}, the present survey limits itself to a logical and probabilistic perspective, which it inherits from StarAI, and to developments in neuro-symbolic computation that are consistent with this perspective.
Furthermore, it  focuses on representative and prototypical systems rather than aiming at completeness (which would not be possible given the page limitations). At the same time, unlike many early approaches to neuro-symbolic computation (see \cite{bader2005dimensions} for an overview), which focused more on modeling issues and principles, we focus on approaches that are also used for learning.

The following sections of the paper each describe a
dimension. We summarize various neuro-symbolic approaches
along these dimensions in Table 1. Furthermore, 
for ease of writing, the table mentions for each system
the key reference (so that we do not always have to repeat these
references).

\section{Directed vs undirected}

Within the graphical model community there is a distinction between the {\em directed} and {\em undirected} graphical models \cite{koller2009probabilistic}, which has led to two distinct types of StarAI systems.
The first generalizes directed models, and resembles Bayesian networks; the second generalizes undirected models like Markov networks or random fields. 
The key difference between the two is that the first class of models indicates a natural direction (sometimes the term ``causal'' is used) between the different random variables, while the second one does not.

In StarAI, the first category includes well-known representations such as plate notation \cite{koller2009probabilistic}, probabilistic relational models (PRMs) \cite{friedman1999learning}, probabilistic logic programs (PLPs) \cite{de2015probabilistic},  and Bayesian logic programs (BLPs) \cite{kersting20071}. 
Today the most typical and popular representatives of this category are the probabilistic (logic) programs. 
The second category includes Markov Logic Networks (MLNs) \cite{richardson2006mln} and Probabilistic Soft Logic (PSL) \cite{bach2017psl}. 
They essentially specify a set of weighted constraints, clauses or formulae.

From a logical perspective, the difference amounts to using a form of definite clauses (as in the programming language Prolog) versus the use of full clausal logic or even first order logic. 
On the one side, a definite clause is an expression of the form $ h \leftarrow b_1 \wedge ... \wedge b_n$ where $h$ and the $b_i$ are logical atoms of the form
$p(t_1, ... , t_m)$, with $p$ being a predicate of arity $m$ and the $t_i$ being terms, that is, constants, variables, or structured terms of the form $f(t_1, ..., t_n)$, where $f$ is a functor and the $t_i$ are again terms. 
On the other side, full clausal logic also allows for formulae of the form $ h_1 \vee ... \vee h_m \leftarrow b_1, ... , b_n$.  

In the first type of rule the direction of the implication indicates, just like the direction of the arrows in a Bayesian network, what can be inferred from what. 
In the second type of rule, this relationship is blurred because of the disjunction in the head of the rule, which allows multiple conclusions for the same premises. This explains why the first type of rule is more directly used for inference, while the second more as a constraint. 
It also reflects the kind of knowledge that the user has about the problem. 
With directed models, one can express that a set of variables has a direct ``causal'' influence on another one,  while with undirected ones one expresses a kind of (soft) constraints on a set of variables, that is, that the variables are related to one another. 

Borrowing this view from StarAI, we can devise a first dimension for neuro-symbolic approaches, which relies entirely on the logical perspective outlined above.

The first category includes \nesy{} systems based on Prolog or Datalog, such as  Neural Theorem Provers (NTPs) \cite{rocktaschel2017ntp}, NLProlog \cite{weber2019nlprolog}, DeepProbLog \cite{manhaeve2018deepproblog} and DiffLog \cite{si2019difflog}.
These systems retain the directed nature of  logical inference as they exploit backward chaining. 
Lifted Relational Neural Networks (LRNNs) \cite{sourek2018lrnn} and $\partial$ILP \cite{evans:dilp} are other examples of non-probabilistic directed models, where  definite clauses are compiled into a neural network architecture in a forward chaining fashion.
The systems that imitate logical reasoning with tensor calculus, Neural Logic Programming (NeuralLP) \cite{Cohen_NeuralLP} and Neural Logic Machines (NLM) \cite{NLM},  are likewise instances of directed logic.

The undirected \nesy{} approaches consider logic as a constraint on the behaviour of a predictive model.
A large group of approaches, including Semantic Based regularization (SBR) \cite{diligenti2017sbr} and Semantic Loss (SL) \cite{xu2018semantic}, exploits logical knowledge as a soft constraint over the hypothesis space in a way that  favours solutions consistent with the encoded knowledge.
SBR implements predicates as neural networks and translates the provided logical formulas into a real valued regularization by means of fuzzy logic, while SL uses marginal probabilities of the target atoms to define the regularization term and relies on  arithmetic circuits \cite{darwiche2011sdd} to evaluate it efficiently.

Another group of approaches, including Logic Tensor Networks (LTN) \cite{donadello2017ltn}, Neural Markov Logic Networks (NMLN) \cite{marra2019nmln} and Relational Neural Machines (RNM) \cite{marra2020relational} extend MLNs,  allowing either predicates (LTN) or factors (NMLN and RNM) to be implemented as neural architectures.
Finally, \cite{rocktaschel2015injecting,demeester2016lifted} compute ground atoms scores as dot products between relation and entities embeddings; implication rules are then translated into a logical loss by means of continuous relaxation of the implication operator. 


\section{Grounding vs proofs}
From a logical perspective there is  a model-theoretic and a proof-theoretic perspective to inference. This is clear when looking at the difference between Answer Set Programming
and the programming language Prolog. In the model theoretic perspective, one first grounds out the clauses in the theory and then calls a SAT solver (possibly after  breaking  cycles), while in a proof theoretic perspective, one performs a sequence of inference steps in order to obtain a proof. 

Grounding is the step whereby  a clause  $c$ (or formula) containing variables $\{V_1, ... , V_k\}$ is replaced by all  instances  $ c\theta$ where $\theta$ is a substitution $\{V_1 = c_1, ... V_k = c_k\}$ and the $c_i$ are constants (or other ground terms) appearing in the domain. The resulting  clause $ c\theta$ is that obtained by simultaneously replacing all variables by the corresponding constants. Usually the grounding process is optimised in order to obtain only those  ground clauses that are relevant for the considered inference task.

These two perspectives carry over to the StarAI perspective.
Many StarAI systems use the logic as a kind of template 
to ground out the relational model in order to obtain 
a grounded model and perform inference. 
This grounded model can be a
graphical model, or alternatively, it can be a ground  weighted logical theory on which traditional inference methods apply, such as belief propagation or weighted model counting.  This is used in well known systems such as MLNs, PSL, BLPs,  and PRMs.  
Some systems like PRMs and BLPs also use aggregates or combining rules in their knowledge base construction approach.  
The idea then is to combine multiple conditional probability distributions into one using, e.g., noisy-or.

Alternatively, one can follow a proof or trace based approach to define the probability distribution and perform inference.  This is akin to what happens in  probabilistic programming  (cf. also \cite{russell2015unifying}), in StarAI frameworks such as PLPs
, probabilistic  databases \cite{van2017query} and  probabilistic unification based grammars such as Stochastic Logic Programs (SLPs) \cite{muggleton1996stochastic}. The idea is that a proof will form the basis for probabilistic inference.  Just like pure logic supports the model-theoretic and proof-theoretic, both perspectives have been explored in parallel for some of the probabilistic logic programming languages  such as ICL \cite{poole2008independent} and ProbLog \cite{fierens2015inference}.

Again this carries over to neuro-symbolic methods.
Approaches of NTPs, DeepProblog, $\partial$ILP and DiffLog are proof-based.
The probabilities or certainties that these systems output are based on the enumerated proofs, and they are also able to learn how to combine them. 
In contrast, approaches of LRNN, LTNs, RNM, NMLN, NLM and NeuralLP are all based on grounding.
Learning in these models is done through learning the (shared) parameters over the ground model and inference is based on possible groundings of the model. 

\section{Logic vs Probability vs Neural}

When two paradigms are integrated, examining which of the base paradigms are preserved, and to which extent, tells us a lot about the strengths and weaknesses of the resulting paradigm.
In StarAI, the traditional knowledge based model construction approach is to use the logic only to generate a probabilistic graphical model, implying that both the inference and the semantics are pushed inside the graphical model. The effect is that it is often harder to reason at a purely logical level with such systems. What is meant here is that it may become unclear how to apply logical inference rules such as resolution (or extensions that take into account the parameters) to such models or what the effect of applying such rules will be.  This is what happens with systems such as PRMs, BLPs, PSL , and MLNs.  
For instance, in MLNs the addition of the resolvent of two weighted rules, makes it hard to predict the effect on the distribution. 
On the other hand, the opposite holds for PLPs and its variants. 
While it is clear what the effect of a logical operation is, it is often harder to directly identify and exploit properties such as conditional or contextual independencies, which are needed for efficient probabilistic inference. 


This position on the spectrum between logic and probability has a profound influence on the properties of the underlying model. 
For \nesy{}, the spectrum involves not only logic and neural networks, but often also probability. 
It has been argued that when combining different perspectives in one model or framework, such as neural, logic and probabilistic ones,  it is  desirable  to have the originals or base paradigms as a special case, see also \cite{de2019neuro}.

The vast majority of current \nesy{} approaches  focus  on the neural aspect (i.e., they originated as a fully neural method to which logical components have been added). Some of these approaches like LTNs and TensorLog \cite{cohen2017tensorlog} pursue a kind of knowledge-based model construction approach in which the logic is compiled away into the neural network architecture.
A different family of \nesy{} approaches, which includes SL and SBR, turns the logic into a regularization function to provide a penalty whenever
the desired logical theory or constraints are violated. 
This leads to the logic being compiled into the weights of the trained neural network.

A small number of NeSy methods, however, retain the focus on logic. 
Some of these methods start from existing logic (programming) frameworks and extend them with primitives that allow them to interface with neural networks and allow for differentiable operations.
Examples include DeepProbLog and  DiffLog.
Other methods instead take an existing framework and turn it into a differentiable version. The key inference concepts are mapped onto an analogous concept that behaves identically for the edge cases, but is continuous and differentiable in non-deterministic cases.
Such methods include $\partial$ILP, $\partial$4 \cite{BosnjakDFI} and NTPs.

Even for methods that focus on logic, it can be useful to map the problem onto an intermediate representation. 
One such idea concerns performing probabilistic inference by mapping it onto a  weighted model counting (WMC) problem. 
This can then in turn be solved by compiling it into a structure (e.g. an arithmetic circuit) that allows for efficient inference.
This has the added benefit that this structure is differentiable, which can facilitate the integration between logic based systems and neural networks. 
DeepProbLog, for example, uses this approach. 
In \cite{pedroassembly}, the authors argue that this intermediate representation can serve as an \emph{assembly language for AI}.

\section{Semantics}
\label{sec:semantics}
Traditionally, StarAI combines two semantics: a logical and a probabilistic one. In a \textit{logical} semantics, atoms are assigned a truth value in the $\{\textit{true}, \textit{false}\}$ set (i.e. $\{0,1\}$). In a \textit{probabilistic} semantics, 
 probability is defined as a measure over sets of possible worlds, where each possible world is an assignment of values to the random variables. 
 This implies that a probabilistic logic semantics defines  probability distributions over ground logical interpretations, that is, over sets of ground facts.
Prominent examples in  StarAI are ProbLog (from the directed side) and Markov Logic (from the undirected one). However, the complexity of inference in probabilistic logic 
has led to statistical relational approaches (e.g. \cite{bach2017psl}), where the truth values are relaxed in the continuous interval $[0,1]$ and logic operators are turned into real valued functions. This setting is described in terms of \textit{fuzzy logic} (or soft logic) semantics, mathematically grounded in the t-norm theory. By exploiting the translation of Boolean formulas into real valued functions, the fuzzy semantics allows to exploit algebraic and geometric properties of t-norms (including especially their differentiability) to reduce complexity. The main issue of fuzzy semantics in the context of StarAI is that it is often not exploited to describe problems that are intrinsically \textit{vague} \cite{fine1975vagueness}, but, simplistically, as a continuous surrogate of Boolean logic. A side effect of this approximation is that many properties of the original logical theory can be realised in many different ways in their continuous translation. Indeed, the \textit{fuzzification} procedure alters  the logical properties of the original theory (such as satisfiability), depending on the particular connectives exploited in the conversion. 
For example, 
in the \textit{\L ukasiewicz t-norm}  $t_{\L}(x,y) = \max\{0, x + y - 1\}$, the conjunction can be 0 (i.e. \textit{false}) even without any of the elements being 0 (e.g. $x=y=0.5$).

Neuro-symbolic approaches can easily be categorized in terms of the  same logical, probabilistic or fuzzy semantics. 
Neural enhancements of the \textit{logic} semantics either use neural networks to turn perceptive input to a logical atom or introduce a relaxed version of logical reasoning performed through tensor calculus.
An instance of the former is ABL \cite{NeSyAbduction}, which use logical abduction to provide the feedback for a neural model processing the perceptive input.
Tensor calculus approaches, such as NLM and NeuralLP, interpret predicates as tensors grounded over all constants in a domain and interpret clauses as a product of those matrices. 

Neural enhancements of the \textit{probabilistic} semantics usually  reparameterize  the underlying distribution in terms of neural components. 
In particular, DeepProbLog exploits neural predicates to compute the probabilities of probabilistic facts as the output of neural computations over vectorial representations of the constants, which is similar to SL in the propositional counterpart. NMLN and RNM use neural potentials in order to implement factors (or their weights) as neural networks. \cite{rocktaschel2015injecting} computes marginal probabilities as logistic functions over similarity measures between embeddings of entities and relations.

Neural enhancements of the \textit{fuzzy} semantics are usually realised by allowing continuous truth values to be the outcome of a neural process and the differentiability of the corresponding t-norm allows for an easy integration with neural computation frameworks. In particular, SBR and LTN turn atoms into neural networks taking as inputs the feature representation of the constants and returning the corresponding truth value.
Similarly, in LRNN and \cite{wang2019integrating}, 
the output of the neurons of the logical network can be interpreted as fuzzy truth values of the corresponding atoms. 

Finally, there is a large class of methods \cite{minervini2017adversarial,demeester2016lifted,cohen2017tensorlog,weber2019nlprolog} realised by relaxing logical statements in a numeric way, without giving any other specific semantics, either probabilistic or fuzzy. Here, atoms are assigned scores in $\mathbb{R}$ computed by a neural scoring function over embeddings. Numerical approximations are then applied  either to combine these scores according to logical formulas or to aggregate proofs scores. The resulting neural architecture is usually differentiable and, thus, trained end-to-end.

\section{Learning parameters or structure}
\label{sec:struct}

StarAI distinguishes between two types of learning: structure learning, which corresponds to learning the logical clauses of the model \cite{KokStruct}, and parameter learning in which the probabilities or weights of the clauses have to be estimated \cite{problogWeights,LowdWeights}.

This distinction is less clear in the \nesy{} setting.
Unlike what is common in StarAI, the \nesy{} approaches do not perform a search through the discrete space of possible clauses,  but rather through the space of parameters of such clauses which are typically enumerated by following a template (often with a predefined complexity).
Examples of such systems include NTPs, $\partial$ILP, DeepProbLog, NeuralLP and DiffLog. 
Alternatively, one can provide a \textit{sketch} of the desired program -- a program with certain decisions left blank --  and learn a \nesy{} model to fill out the blanks, such as DeepProbLog and $\partial$4.

A substantial number of approaches tries to leverage the best of both worlds.
These ideas include using neural models to guide the symbolic search \cite{NGSynth,ellis:libraries,Valkov2018HOUDINILL}, or using a neural model to produce a program that is then executed symbolically \cite{ellis:images,mao2018the}.


\section{Symbols vs Sub-symbols}

An important factor in both StarAI and \nesy{} systems is the representation of entities. 
StarAI generally represents entities by  constants (symbols). 
But neural methods are  numerical by nature and therefore symbols are replaced with sub-symbols, i.e., vectorized representations.
If the entity has inherent numerical properties, these could be used as sub-symbols (e.g. the pixel data of an image). 
However, if this is not the case, a one-hot encoding or learned embedding can be used instead. 
This, of course, has an impact on the generalizability of the system towards unseen entities, as  new embeddings have to be learned for new symbols.
Naturally, among the neuro-symbolic methods, there is a wide variety in how symbols and sub-symbols are used in representation and reasoning.
The idea of mapping entities onto sub-symbols is made very explicit in LTNs, where  in a first step, all symbols are replaced with sub-symbols.
In DeepProbLog, entities are represented using symbols, but they sometimes have sub-symbolic representations that are only used inside the neural networks.
Similarly, in \cite{lippi2009prediction} and RNM, MLNs are conditioned on a feature representation of constants (e.g. images, audio signals, etc.). Finally, among those models exploiting learned embeddings, we find \cite{rocktaschel2015injecting,minervini2017adversarial,demeester2016lifted}.

Now that we discussed how entities can be represented by symbols and sub-symbols, let us discuss how they can be used for reasoning. 
Most methods either only work with logic reasoning on symbols, or perform algebraic operations on sub-symbols. However, some methods can use both simultaneously.
A very powerful and elegant mechanism for reasoning about symbols in first order logic is  {\em unification}. It is used to reason about  equality at the symbolic level. For instance,  the atomic expressions $p(a,Y)$ and $p(X,b)$ can be unified using the substitution $\{X=a, Y=b\}$. Unification not only works for constants but also for structured terms $f(t_1, ... ,t_n)$ where $f$ is a structured term and the $t_i$ are constants, variables or structured terms themselves. 

While unification is not supported by standard neural networks,
reasoning about equality corresponds  closely to 
reasoning about similarity in embedding space.  Entities 
are typically embedded in some metric space, and represented 
through their embeddings, that is, through sub-symbols.
Reasoning typically proceeds by performing algebraic operations (such as vector addition) on these embeddings, and considering the similarity 
between two entities by using their distance in embedding space. 
It is quite interesting to see to what extent current neuro-symbolic approaches support unification on the one hand, and to what extent the use of embeddings has been integrated into the neuro-symbolic logics as a kind of {\em soft} equality or unification 

This idea was implemented in NTPs and NLProlog as \emph{soft} or \emph{weak unification}. In these systems, two entities can be unified if they are similar, and not just if they are identical. As such, this system can interweave both symbols and sub-symbols during inference.
For each entity, an embedding is learned and their similarity is determined based on the distance between the embeddings using a radial basis function. However, this potentially adds a lot of different proof paths, which can result in computational issues for larger programs. This problem was solved in later iterations of the system \cite{minervini2019differentiable}. 

\section{Type of logic}

There is a natural ordering of logical representations,
starting with propositional logic (only arity 0 predicates), to relational
logic (having no structured terms, so only constants and variables as terms, which is also the basis for the Datalog database language), to general first order logic (FOL), 
and then to logic programs (LP) as in the programming language Prolog.  Logic programs are usually restricted to definite clauses, while the semantics of a definite clause program is given by its least Herbrand model, the set of all ground facts that are logically entailed by the program.  This contrasts with the standard
semantics of first order logic that would also allow for other models.
This difference carries over to StarAI, where
probabilistic logic programs and Markov Logic 
inherit their semantics from logic programming, respectively
first order logic. This explains, for instance, why
Markov Logic's semantics boils down to a maximum entropy approach when a theory has multiple models (such as $a \vee b$), cf. \cite{de2015probabilistic,raedt2016statistical} for more details.
On the other hand, logic programs are also the basis for the programming language Prolog, which implies that they can be used to specify traditional programs such as sorting and data structures such as lists through structured terms. This is relevant especially for those approaches to neurosymbolic computation that are used to synthesize programs from examples.


Neuro-symbolic representations typically extend one of these four types of logic: propositional, relational, first order logic, or logic programs. 
For instance, SL focuses only on the propositional setting.  On the other hand,  $\partial$ILP, NTPs and DiffLog are based on Datalog, which belongs to relational logic segment. LTNs and SBR use fuzzy logic to translate a general FOL theory into a training objective, either isolated or in conjunction with a supervised criterion.  Just like Markov Logic, also RNM and NMLN use first order logic to generate a random field. Finally, DeepProbLog, NLProlog and LRNN are examples of neuro-symbolic logic programming frameworks.

\begin{table*}[t]
\setlength{\tabcolsep}{2.8pt}
\renewcommand{\arraystretch}{1.3}
    \centering
    \resizebox{\textwidth}{!}{
    \begin{tabular}{lccccccc}
        \toprule
         & \textbf{Dimension 1} &  \textbf{Dimension 2} &  \textbf{Dimension 3} &  \textbf{Dimension 4} &  \textbf{Dimension 5} &  \textbf{Dimension 6} &  \textbf{Dimension 7} \\
         
         &  \makecell[l]{(D)irected \\ (U)ndirected} & \makecell[l]{(G)rounding \\ (P)roofs} & \makecell[l]{(L)ogic  \\ (P)robability \\ (N)eural}  & \makecell[l]{(L)ogic \\ (P)robability \\ (F)uzzy} & \makecell[l]{(P)arameter \\ (S)tructure}  & \makecell[l]{(S)ymbols \\ (Sub)symbols} & \makecell[l]{(P)ropositional \\ (R)elational \\ (FOL) \\ (LP)}\\
         \midrule
         $\partial$ILP  \cite{evans:dilp} & D & P & L+N & L & P & S & R \\
         DeepProbLog \cite{manhaeve2018deepproblog} & D & P & L+P+N & P & P & S+Sub & LP \\ 
         DiffLog \cite{si2019difflog} & D & P & L+N & L & P+S & S & R \\
         LRNN \cite{sourek2018lrnn}& D & P & L+N & F & P+S & S+Sub & LP \\
         LTN \cite{donadello2017ltn}& U & G & L+N & F & P & Sub & FOL \\
         NeuralLP \cite{Cohen_NeuralLP} & D & G & L+N & L & P & S & R \\
         NLM  \cite{NLM} & D & G & L+N & L & P+S & S & R \\
         NLProlog \cite{weber2019nlprolog}& D & P & L+P+N & P & P+S & S+Sub & LP\\
         NMLN \cite{marra2019nmln} & U & G & L+P+N & P & P+S & S+Sub & FOL \\
         NTP \cite{rocktaschel2017ntp}& D & P & L+N & L & P+S & S+Sub & R \\
         RNM \cite{marra2020relational}& U & G & L+P+N & P & P & S+Sub & FOL \\
         SL \cite{xu2018semantic}
         & U & G & L+P+N & P & P & S+Sub & P \\
         SBR \cite{diligenti2017sbr} & U & G & L+N & F & P & Sub & FOL \\
         Tensorlog \cite{cohen2017tensorlog} & D & P & L+N & P & P & S+Sub & R\\
         \bottomrule

    \end{tabular}}
    \caption{Taxonomy of a (non-exhaustive) list of NeSy models according to the 7 dimensions outlined in the paper.}
    \label{tab:my_label}
\end{table*}

\section{Open challenges}
To conclude, we now list a number of challenges for \nesy{}, which deserve, in our opinion, more attention.

\paragraph{Probabilistic reasoning}
Although relatively few methods explore the integration of  logical and neural methods through probabilities perspective, we believe that a probabilistic approach is the best way to principally integrate the two \cite{de2019neuro}.
There should be further investigation into the applicability of probabilistic reasoning for neuro-symbolic computation.

\paragraph{Structure learning}
While significant progress has been made on learning the structure of purely relational models (without probabilities), learning StarAI models remains a major challenge due to the complexity of inference and the combinatorial nature of the problem.
Incorporating neural aspects complicates the problem even more.
\nesy{} methods have certainly shown potential for addressing this problem (Section \ref{sec:struct}), but the existing methods are still limited and mostly domain-specific which impedes their wide application.

\paragraph{Scaling inference}
Scalable inference is a major challenge for StarAI and therefore also for \nesy{} approaches with an explicit logical or probabilistic reasoning component.
Investigating to which extent neural methods can help with this challenge by means of lifted (exploiting symmetries in models) or approximate inference, as well as reasoning from the intermediate representations \cite{ACL-AAAI20}, are promising future research directions.

\paragraph{Data efficiency}
A major advantage of StarAI methods, as compared to neural ones, is their data efficiency -- StarAI methods can efficiently learn from small amount of data, whereas neural methods are data hungry.
On the other hand, StarAI methods do not scale to big data sets, while neural methods can easily handle them.
We believe that understanding how these methods can help each other to overcome their complementary weaknesses, is a promising research direction.

\paragraph{Symbolic representation learning}
The effectiveness of deep learning  comes from the ability to change the representation of the data so that the target task becomes easier to solve.
The ability to change the representation on the symbolic level as well would significantly increase the capabilities of \nesy{} systems.
This is a major open challenge for which neurally inspired methods could help achieve progress \cite{crop:playgol,dumancic:encoding}.

\section*{Acknowledgements}
Robin Manhaeve and Sebastijan Dumancic are funded by the Research Foundation-Flanders (FWO).
This work has also received funding from the European Research Council (ERC) under the European Union’s Horizon 2020 research and innovation programme (grant agreement No [694980] SYNTH: Synthesising Inductive Data Models).

{
\small
\bibliographystyle{named}
\bibliography{ourbib15}

\begin{thebibliography}{}

\bibitem[\protect\citeauthoryear{Abboud \bgroup \em et al.\egroup
  }{2020}]{ACL-AAAI20}
Ralph Abboud, {\.I}smail~{\.I}lkan Ceylan, and Thomas Lukasiewicz.
\newblock Learning to reason: Leveraging neural networks for approximate {DNF}
  counting.
\newblock In {\em AAAI}, 2020.

\bibitem[\protect\citeauthoryear{Bach \bgroup \em et al.\egroup
  }{2017}]{bach2017psl}
Stephen~H. Bach, Matthias Broecheler, Bert Huang, and Lise Getoor.
\newblock Hinge-loss markov random fields and probabilistic soft logic.
\newblock {\em J. Mach. Learn. Res.}, 18:109:1--109:67, 2017.

\bibitem[\protect\citeauthoryear{Bader and Hitzler}{2005}]{bader2005dimensions}
Sebastian Bader and Pascal Hitzler.
\newblock Dimensions of neural-symbolic integration-a structured survey.
\newblock {\em arXiv preprint cs/0511042}, 2005.

\bibitem[\protect\citeauthoryear{Besold \bgroup \em et al.\egroup
  }{2017}]{besold2017neural}
Tarek~R Besold, Artur~d'Avila Garcez, Sebastian Bader, Howard Bowman, Pedro
  Domingos, Pascal Hitzler, Kai-Uwe K{\"u}hnberger, Luis~C Lamb, Daniel Lowd,
  Priscila Machado~Vieira Lima, et~al.
\newblock Neural-symbolic learning and reasoning: A survey and interpretation.
\newblock {\em arXiv preprint arXiv:1711.03902}, 2017.

\bibitem[\protect\citeauthoryear{Bo\v{s}njak \bgroup \em et al.\egroup
  }{2017}]{BosnjakDFI}
Matko Bo\v{s}njak, Tim Rockt\"{a}schel, Jason Naradowsky, and Sebastian Riedel.
\newblock Programming with a differentiable forth interpreter.
\newblock In {\em ICML}, 2017.

\bibitem[\protect\citeauthoryear{Cohen \bgroup \em et al.\egroup
  }{2017}]{cohen2017tensorlog}
William~W. Cohen, Fan Yang, and Kathryn Mazaitis.
\newblock Tensorlog: Deep learning meets probabilistic dbs.
\newblock {\em CoRR}, abs/1707.05390, 2017.

\bibitem[\protect\citeauthoryear{Cropper}{2019}]{crop:playgol}
Andrew Cropper.
\newblock Playgol: Learning programs through play.
\newblock In {\em {IJCAI} 2019}, 2019.

\bibitem[\protect\citeauthoryear{Dai \bgroup \em et al.\egroup
  }{2019}]{NeSyAbduction}
Wang-Zhou Dai, Qiuling Xu, Yang Yu, and Zhi-Hua Zhou.
\newblock Bridging machine learning and logical reasoning by abductive
  learning.
\newblock In {\em NeurIPS}. 2019.

\bibitem[\protect\citeauthoryear{Darwiche}{2011}]{darwiche2011sdd}
Adnan Darwiche.
\newblock Sdd: A new canonical representation of propositional knowledge bases.
\newblock In {\em {IJCAI}}, 2011.

\bibitem[\protect\citeauthoryear{{De Raedt} and
  Kimmig}{2015}]{de2015probabilistic}
Luc {De Raedt} and Angelika Kimmig.
\newblock Probabilistic (logic) programming concepts.
\newblock {\em Machine Learning}, 100(1):5--47, 2015.

\bibitem[\protect\citeauthoryear{{De Raedt} \bgroup \em et al.\egroup
  }{2016}]{raedt2016statistical}
Luc {De Raedt}, Kristian Kersting, Sriraam Natarajan, and David Poole.
\newblock Statistical relational artificial intelligence: Logic, probability,
  and computation.
\newblock {\em Synthesis Lectures on Artificial Intelligence and Machine
  Learning}, 10(2):1--189, 2016.

\bibitem[\protect\citeauthoryear{{De Raedt} \bgroup \em et al.\egroup
  }{2019}]{de2019neuro}
Luc {De Raedt}, Robin Manhaeve, Sebastijan Dumančić, Thomas Demeester, and
  Angelika Kimmig.
\newblock Neuro-symbolic= neural+ logical+ probabilistic.
\newblock In {\em NeSy @ IJCAI}, 2019.

\bibitem[\protect\citeauthoryear{Demeester \bgroup \em et al.\egroup
  }{2016}]{demeester2016lifted}
Thomas Demeester, Tim Rockt{\"{a}}schel, and Sebastian Riedel.
\newblock Lifted rule injection for relation embeddings.
\newblock In {\em {EMNLP}}, 2016.

\bibitem[\protect\citeauthoryear{Diligenti \bgroup \em et al.\egroup
  }{2017}]{diligenti2017sbr}
Michelangelo Diligenti, Marco Gori, and Claudio Sacc{\`{a}}.
\newblock Semantic-based regularization for learning and inference.
\newblock {\em Artif. Intell.}, 2017.

\bibitem[\protect\citeauthoryear{Donadello \bgroup \em et al.\egroup
  }{2017}]{donadello2017ltn}
Ivan Donadello, Luciano Serafini, and Artur~S. d'Avila Garcez.
\newblock Logic tensor networks for semantic image interpretation.
\newblock In Carles Sierra, editor, {\em {IJCAI}}, 2017.

\bibitem[\protect\citeauthoryear{Dong \bgroup \em et al.\egroup }{2019}]{NLM}
Honghua Dong, Jiayuan Mao, Tian Lin, Chong Wang, Lihong Li, and Denny Zhou.
\newblock Neural logic machines.
\newblock In {\em ICLR}, 2019.

\bibitem[\protect\citeauthoryear{Dumančić \bgroup \em et al.\egroup
  }{2019}]{dumancic:encoding}
Sebastijan Dumančić, Tias Guns, Wannes Meert, and Hendrik Blockeel.
\newblock Learning relational representations with auto-encoding logic
  programs.
\newblock In {\em {IJCAI} 2019}, 2019.

\bibitem[\protect\citeauthoryear{Ellis \bgroup \em et al.\egroup
  }{2018a}]{ellis:libraries}
Kevin Ellis, Lucas Morales, Mathias Sabl\'{e}-Meyer, Armando Solar-Lezama, and
  Josh Tenenbaum.
\newblock Learning libraries of subroutines for neurally\textendash guided
  bayesian program induction.
\newblock In {\em NeurIPS}. 2018.

\bibitem[\protect\citeauthoryear{Ellis \bgroup \em et al.\egroup
  }{2018b}]{ellis:images}
Kevin Ellis, Daniel Ritchie, Armando Solar{-}Lezama, and Josh Tenenbaum.
\newblock Learning to infer graphics programs from hand-drawn images.
\newblock In {\em NeurIPS 2018}, pages 6062--6071, 2018.

\bibitem[\protect\citeauthoryear{Evans and Grefenstette}{2018}]{evans:dilp}
Richard Evans and Edward Grefenstette.
\newblock Learning explanatory rules from noisy data.
\newblock {\em J. Artif. Intell. Res.}, 61:1--64, 2018.

\bibitem[\protect\citeauthoryear{Fierens \bgroup \em et al.\egroup
  }{2015}]{fierens2015inference}
Daan Fierens, Guy Van~den Broeck, Joris Renkens, Dimitar Shterionov, Bernd
  Gutmann, Ingo Thon, Gerda Janssens, and Luc De~Raedt.
\newblock Inference and learning in probabilistic logic programs using weighted
  boolean formulas.
\newblock {\em Theory and Practice of Logic Programming}, 15(3):358--401, 2015.

\bibitem[\protect\citeauthoryear{Fine}{1975}]{fine1975vagueness}
Kit Fine.
\newblock Vagueness, truth and logic.
\newblock {\em Synthese}, pages 265--300, 1975.

\bibitem[\protect\citeauthoryear{Friedman \bgroup \em et al.\egroup
  }{1999}]{friedman1999learning}
Nir Friedman, Lise Getoor, Daphne Koller, and Avi Pfeffer.
\newblock Learning probabilistic relational models.
\newblock In {\em IJCAI}, volume~99, pages 1300--1309, 1999.

\bibitem[\protect\citeauthoryear{Garcez \bgroup \em et al.\egroup
  }{2019}]{garcez2019neural}
Artur~d'Avila Garcez, Marco Gori, Luis~C Lamb, Luciano Serafini, Michael
  Spranger, and Son~N Tran.
\newblock Neural-symbolic computing: An effective methodology for principled
  integration of machine learning and reasoning.
\newblock {\em arXiv preprint arXiv:1905.06088}, 2019.

\bibitem[\protect\citeauthoryear{Getoor and Taskar}{2007}]{Getoor07:book}
L.~Getoor and B.~Taskar, editors.
\newblock {\em An Introduction to Statistical Relational Learning}.
\newblock {MIT} {P}ress, 2007.

\bibitem[\protect\citeauthoryear{Gutmann \bgroup \em et al.\egroup
  }{2008}]{problogWeights}
Bernd Gutmann, Angelika Kimmig, Kristian Kersting, and Luc De~Raedt.
\newblock Parameter learning in probabilistic databases: A least squares
  approach.
\newblock In {\em ECML\&PKDD}, 2008.

\bibitem[\protect\citeauthoryear{Kalyan \bgroup \em et al.\egroup
  }{2018}]{NGSynth}
Ashwin Kalyan, Abhishek Mohta, Oleksandr Polozov, Dhruv Batra, Prateek Jain,
  and Sumit Gulwani.
\newblock Neural-guided deductive search for real-time program synthesis from
  examples.
\newblock In {\em {ICLR}}, 2018.

\bibitem[\protect\citeauthoryear{Kersting and De~Raedt}{2007}]{kersting20071}
Kristian Kersting and Luc De~Raedt.
\newblock Bayesian logic programming: Theory and tool.
\newblock {\em Statistical Relational Learning}, page 291, 2007.

\bibitem[\protect\citeauthoryear{Kok and Domingos}{2005}]{KokStruct}
Stanley Kok and Pedro Domingos.
\newblock Learning the structure of markov logic networks.
\newblock ICML, 2005.

\bibitem[\protect\citeauthoryear{Koller and
  Friedman}{2009}]{koller2009probabilistic}
Daphne Koller and Nir Friedman.
\newblock {\em Probabilistic Graphical Models - Principles and Techniques}.
\newblock {MIT} Press, 2009.

\bibitem[\protect\citeauthoryear{Lippi and
  Frasconi}{2009}]{lippi2009prediction}
Marco Lippi and Paolo Frasconi.
\newblock Prediction of protein $\beta$-residue contacts by markov logic
  networks with grounding-specific weights.
\newblock {\em Bioinformatics}, 25(18):2326--2333, 2009.

\bibitem[\protect\citeauthoryear{Lowd and Domingos}{2007}]{LowdWeights}
Daniel Lowd and Pedro Domingos.
\newblock Efficient weight learning for markov logic networks.
\newblock ECMLPKDD, 2007.

\bibitem[\protect\citeauthoryear{Manhaeve \bgroup \em et al.\egroup
  }{2018}]{manhaeve2018deepproblog}
Robin Manhaeve, Sebastijan Dumančić, Angelika Kimmig, Thomas Demeester, and
  Luc~De Raedt.
\newblock Deepproblog: Neural probabilistic logic programming.
\newblock In {\em NeurIPS}, 2018.

\bibitem[\protect\citeauthoryear{Mao \bgroup \em et al.\egroup
  }{2019}]{mao2018the}
Jiayuan Mao, Chuang Gan, Pushmeet Kohli, Joshua~B. Tenenbaum, and Jiajun Wu.
\newblock The neuro-symbolic concept learner: Interpreting scenes, words, and
  sentences from natural supervision.
\newblock In {\em ICLR}, 2019.

\bibitem[\protect\citeauthoryear{Marra and Kuželka}{2019}]{marra2019nmln}
Giuseppe Marra and Ondrej Kuželka.
\newblock Neural markov logic networks.
\newblock {\em CoRR}, abs/1905.13462, 2019.

\bibitem[\protect\citeauthoryear{Marra \bgroup \em et al.\egroup
  }{2020}]{marra2020relational}
Giuseppe Marra, Michelangelo Diligenti, Francesco Giannini, Marco Gori, and
  Marco Maggini.
\newblock Relational neural machines.
\newblock In {\em {ECAI}}, 2020.

\bibitem[\protect\citeauthoryear{Minervini \bgroup \em et al.\egroup
  }{2017}]{minervini2017adversarial}
Pasquale Minervini, Thomas Demeester, Tim Rockt{\"{a}}schel, and Sebastian
  Riedel.
\newblock Adversarial sets for regularising neural link predictors.
\newblock In {\em {UAI}}, 2017.

\bibitem[\protect\citeauthoryear{Minervini \bgroup \em et al.\egroup
  }{2020}]{minervini2019differentiable}
Pasquale Minervini, Matko Bo{\v{s}}njak, Tim Rockt{\"a}schel, Sebastian Riedel,
  and Edward Grefenstette.
\newblock Differentiable reasoning on large knowledge bases and natural
  language.
\newblock In {\em AAAI}, 2020.

\bibitem[\protect\citeauthoryear{Muggleton}{1996}]{muggleton1996stochastic}
Stephen Muggleton.
\newblock Stochastic logic programs.
\newblock {\em Advances in inductive logic programming}, 32:254--264, 1996.

\bibitem[\protect\citeauthoryear{Poole}{2008}]{poole2008independent}
David Poole.
\newblock The independent choice logic and beyond.
\newblock In {\em Probabilistic inductive logic programming}, pages 222--243.
  Springer, 2008.

\bibitem[\protect\citeauthoryear{Richardson and
  Domingos}{2006}]{richardson2006mln}
Matthew Richardson and Pedro~M. Domingos.
\newblock Markov logic networks.
\newblock {\em Machine Learning}, 62(1-2):107--136, 2006.

\bibitem[\protect\citeauthoryear{Rockt{\"{a}}schel and
  Riedel}{2017}]{rocktaschel2017ntp}
Tim Rockt{\"{a}}schel and Sebastian Riedel.
\newblock End-to-end differentiable proving.
\newblock In {\em NIPS}, 2017.

\bibitem[\protect\citeauthoryear{Rockt{\"{a}}schel \bgroup \em et al.\egroup
  }{2015}]{rocktaschel2015injecting}
Tim Rockt{\"{a}}schel, Sameer Singh, and Sebastian Riedel.
\newblock Injecting logical background knowledge into embeddings for relation
  extraction.
\newblock In {\em {NAACL} {HLT}}, 2015.

\bibitem[\protect\citeauthoryear{Russell}{2015}]{russell2015unifying}
Stuart Russell.
\newblock Unifying logic and probability.
\newblock {\em Communications of the ACM}, 58(7):88--97, 2015.

\bibitem[\protect\citeauthoryear{Si \bgroup \em et al.\egroup
  }{2019}]{si2019difflog}
Xujie Si, Mukund Raghothaman, Kihong Heo, and Mayur Naik.
\newblock Synthesizing datalog programs using numerical relaxation.
\newblock In {\em {IJCAI}}, 2019.

\bibitem[\protect\citeauthoryear{Valkov \bgroup \em et al.\egroup
  }{2018}]{Valkov2018HOUDINILL}
Lazar Valkov, Dipak Chaudhari, Akash Srivastava, Charles~A. Sutton, and Swarat
  Chaudhuri.
\newblock Houdini: Lifelong learning as program synthesis.
\newblock In {\em NeurIPS}, 2018.

\bibitem[\protect\citeauthoryear{Van~den Broeck \bgroup \em et al.\egroup
  }{2017}]{van2017query}
Guy Van~den Broeck, Dan Suciu, et~al.
\newblock Query processing on probabilistic data: A survey.
\newblock {\em Foundations and Trends{\textregistered} in Databases},
  7(3-4):197--341, 2017.

\bibitem[\protect\citeauthoryear{Wang and Pan}{2019}]{wang2019integrating}
Wenya Wang and Sinno~Jialin Pan.
\newblock Integrating deep learning with logic fusion for information
  extraction.
\newblock {\em CoRR}, abs/1912.03041, 2019.

\bibitem[\protect\citeauthoryear{Weber \bgroup \em et al.\egroup
  }{2019}]{weber2019nlprolog}
Leon Weber, Pasquale Minervini, Jannes M{\"{u}}nchmeyer, Ulf Leser, and Tim
  Rockt{\"{a}}schel.
\newblock Nlprolog: Reasoning with weak unification for question answering in
  natural language.
\newblock In {\em {ACL}}, 2019.

\bibitem[\protect\citeauthoryear{Xu \bgroup \em et al.\egroup
  }{2018}]{xu2018semantic}
Jingyi Xu, Zilu Zhang, Tal Friedman, Yitao Liang, and Guy~Van den Broeck.
\newblock A semantic loss function for deep learning with symbolic knowledge.
\newblock In {\em {ICML}}, 2018.

\bibitem[\protect\citeauthoryear{Yang \bgroup \em et al.\egroup
  }{2017}]{Cohen_NeuralLP}
Fan Yang, Zhilin Yang, and William~W Cohen.
\newblock Differentiable learning of logical rules for knowledge base
  reasoning.
\newblock In {\em {NIPS} 2017}. 2017.

\bibitem[\protect\citeauthoryear{Zuidberg Dos~Martires \bgroup \em et
  al.\egroup }{2019}]{pedroassembly}
Pedro Zuidberg Dos~Martires, Vincent Derkinderen, Robin Manhaeve, Wannes Meert,
  Angelika Kimmig, and Luc De~Raedt.
\newblock Transforming probabilistic programs into algebraic circuits for
  inference and learning.
\newblock 2019.

\bibitem[\protect\citeauthoryear{Šourek \bgroup \em et al.\egroup
  }{2018}]{sourek2018lrnn}
Gustav Šourek, Vojtech Aschenbrenner, Filip Zelezn{\'{y}}, Steven Schockaert,
  and Ondrej Kuželka.
\newblock Lifted relational neural networks: Efficient learning of latent
  relational structures.
\newblock {\em J. Artif. Intell. Res.}, 62:69--100, 2018.

\end{thebibliography}
}
\end{document}